%% file: main.tex
\documentclass{article}
\usepackage{amsmath,graphicx}
\usepackage{stfloats}
\usepackage{amssymb}
\usepackage{nicematrix,tikz}
\usepackage{algpseudocode}
\usepackage{algorithm}
\usepackage{caption}
\usepackage{natbib}

\DeclareMathOperator*{\argmin}{arg\,min}

\title{Partial Rewriting for Multi-Stage ASR}

\author{Antoine Bruguier, David Qiu, Yanzhang He \\ Google LLC, USA}

\date{\{tonybruguier,qdavid,yanzhanghe\}@google.com}

\begin{document}

\maketitle

\begin{abstract}
\input{abstract}
\end{abstract}

\section{Introduction}
\input{intro}

\section{Goals and constraints}

\input{goal}

\section{Proposed algorithm}

\input{algo}

\section{Results}

\input{results}

\section{Conclusions}
\input{conclusion}

\bibliographystyle{unsrtnat}
\bibliography{mybib}

\end{document}

%% file: abstract.tex
For many streaming automatic speech recognition tasks, it is important to provide timely intermediate streaming results, while refining a high quality final result. This can be done using a multi-stage architecture, where a small left-context only model creates streaming results and a larger left- and right-context model produces a final result at the end. While this significantly improves the quality of the final results without compromising the streaming emission latency of the system, streaming results do not benefit from the quality improvements. Here, we propose using a text manipulation algorithm that merges the streaming outputs of both models. We improve the quality of streaming results by around 10\%, without altering the final results. Our approach introduces no additional latency and reduces flickering. It is also lightweight, does not require retraining the model, and it can be applied to a wide variety of multi-stage architectures.

%% file: intro.tex
\label{sec:intro}

Streaming automatic speech recognition (ASR) processes the audio as it is being spoken, without waiting for the entire utterance to be finished before generating a transcript. The intermediate results are called \textit{partial} results, while the single result produced at the end of the utterance is called a \textit{final} result. This is in contrast to batch ASR which waits until the entire utterance is spoken before generating only a final result. Streaming ASR is useful in applications where it is important to provide feedback to the user as soon as possible, such as voice-activated assistants, dictation, or live video captioning.

The first streaming ASR systems were developed several decades ago~\cite{asr_textbook,conventional,brown_partial_results,selfridge_incremental,incremental_results_multimodal,tv_real_time}, but more recently new approaches relying on end-to-end models~\cite{ms_e2e,fb_e2e,google_e2e,keyboard_dictation_model} have been developed. However, batch ASR usually outperforms streaming ASR because it can better understand the context of the speech and therefore make more accurate predictions.

Cascaded recognizers~\cite{cascaded_e2e_original,cascaded_e2e} attempt to have the best of both worlds by combining the strengths of both streaming and batch ASR systems. They do so by using two models. A \textit{causal} model only uses the left context outputs and therefore has low latency. A \textit{cascaded} model with both left and right contexts looks ahead in the audio to output results and therefore has better accuracy. The cascaded model is also more accurate because it has more weights than the causal one. While both models can generate partial and final results, typically only the causal model is used to generate the partial results so that the latency is low, and only the cascaded model is used to generate the final result so that it is more accurate~\cite{cascaded_latency}. This approach enables low-latency streaming ASR while achieving a high quality for the final results. However, this leaves a gap in quality because the partial results still have relatively lower quality.

Further, running two models independently can cause issues because even though the models decode the same audio and even share some of their weights, there is no coupling between the two outputs and the partial results may occasionally differ substantially from the final result. Recent improvements have been able to tie in several models together~\cite{jay2beamsearches,fast_slow_encoder}. By running the two models together, they are able to improve the partial results, but this comes at the cost of not having a right context for high-quality final results. In essence, both models are causal, but one is still larger than the other. Further, the two models now have to be run synchronously, and the overall architecture is fixed at training time.

In parallel to these modelling efforts, there have been some improvements in how partial results can be evaluated along three dimensions: flickering, partial quality, and partial latency. Flickering occurs when words that have already been decoded in a previous partial results are changed in subsequent partial results. This negatively impacts the experience for the users~\cite{accessibility_ux,ux_study} as well as potentially increasing the computation cost of other systems (e.g. translation~\cite{translation2} or assistants~\cite{result_prefetching}). The amount of flickering can be measured using the metric defined in~\cite{shangguan2020analyzing}. Work by~\cite{deflicker} defines metrics for partial quality and latency. This allows us to measure improvements to partial results along these three dimensions. We will describe the three metrics of interest in more detail in section~\ref{sec:metrics_of_interest}.

%% file: goal.tex
\label{sec:goal}

\subsection{Overview}

In the present work, we want to improve the quality of the partial results by using a large model that has a right context, but we want to do so without scarifying latency. Contrary to~\cite{cascaded_e2e_original,cascaded_e2e} we no longer see partial results as a by-product of the decoding.

Instead, we want to keep two models running completely independently as in~\cite{cascaded_e2e_original} and still allow the larger model to have a right context. We also do not require a fixed architecture at training time. Our work will leave the final results unchanged as they already benefit from the larger model and the use of a right context, but it will improve the quality of partial results significantly.

We want the approach to be cheap computationally. It should not increase the memory budget in any measurable way, nor increase power consumption.

\subsection{Combining results from two models}
The only assumption we make is that we have two models that output a stream of partial results. The results of the first model have low latency, but relatively lower quality. The results from the second model have much higher latency, but are of better quality. Like~\cite{cascaded_e2e_original,cascaded_e2e,cascaded_latency} we want the best of both world, but this time for partial results without tying in models architectures as done in~\cite{jay2beamsearches,fast_slow_encoder} nor require retraining.

In order to validate our approach experimentally, we modify how the cascaded architecture creates partials. While we still only use the cascaded model to create final results, we now allow for \textit{both} causal and cascaded model to create partial results. Then, our proposed approach combines these two streams of partial results into a single stream of new partial results that still have low latency, but better quality. We do not rely on the specificity of the cascaded architecture and \textit{only} assume that we have two models running concurrently and decoding the same audio. We only modify the partial results, and leave the final results unchanged.

\subsection{Metrics of interest}
\label{sec:metrics_of_interest}

In order to measure the validity of our approach, we must be able to measure the quality of our newly created partial results. For this we heavily rely on a \textit{partial word error rate} (PWER) metric defined in section 2.2 of~\cite{deflicker}. Roughly speaking, PWER is the WER averaged over all the partials, ignoring the missing words at the end of each partial.

The causal model outputs partial results with a very low latency, and we do not wish to introduce any new latency. We thus rely on a metric that estimates the latency of \textit{all} partials, as defined in section 2.3 of~\cite{deflicker}. Roughly speaking, the \textit{partial latency} (PL) metric is the average appearance time of every correct word since the beginning of the utterance, where only the change between experiments is the meaningful number.

Finally, an algorithm could increase the amount of flickering of the partial results. In order to control for this potential negative change, we measure the \textit{unstable partial word ratio} (UPWR) as defined in~\cite{shangguan2020analyzing}. Roughly speaking, UPWR is the
fraction of already decoded words in all the partial results that are changed by a subsequent partial or final result. However, in order to measure the effect of flickering more accurately, we expand the definition of the metric by measuring UPWR on different set of results. If the decoding of an utterance produces $N$ partial results and 1 final result, we measure UPWR on three different sets: 1) Only the $N$ partial results, 2) Only the single $N^{\text{th}}$ partial result and the final result, and 3) The $N$ partial results followed by the final result. This allows us to measure how much flickering occurs during the streaming decoding, during the transition from streaming decoding to final result, and overall, respectively.

%% file: algo.tex
\label{sec:algo}

\subsection{Intuition and core algorithm}
\label{sec:intuition}

The algorithm we propose relies on text manipulation only. We take as input the two streams of partial results, one from the causal model and one from the cascaded model, and then create a new stream of composite partial results.

\input{fig_lev}

Consider the example where in the middle of decoding of an utterance, the user has said ``Rosalie how are you''. The causal model, being smaller and less accurate, has incorrectly decoded it as ``\texttt{\textcolor{blue}{\_ro za ee \_how \_are \_you}}'' ($n=6$ word pieces\footnote{We use word-pieces~\cite{google_e2e} to decompose text into tokens even but the approach can also use Unicode codepoints as tokens}), while the cascaded is more accurate but has a delay, and thus has only decoded ``\texttt{\textcolor{purple}{\_ro sa l ie \_how}}'' ($m=5$ word pieces) at the same frame. As expected $m < n$ since the cascaded model uses a right context, but it is not true that the $n - m = 1$ word piece ``\texttt{\textcolor{blue}{\_you}}'' corresponds to amount of audio in the right context. Indeed, if we were to naively copy the first $m$ word pieces from the cascaded model and then append the $n - m = 1$ last word pieces from the causal model, we would get an incorrect transcript: ``\texttt{\textcolor{purple}{\_ro sa l ie \_how} \textcolor{blue}{\_you}}'' because the word ``\texttt{\_are}'' is missing.

\input{algo1}
\input{algo2}

Instead, we need a way to blend the two partial results correctly. Every time we merge two partial results, we should correctly estimate how many extra tokens to copy from the causal results.

We propose to use a Levenshein alignment~\cite{cormen01introduction} of the two sequences, as shown in figure~\ref{fig:lev}. By aligning the two sequences, we are able to compute the alignment costs. We then modify the Levenshtein alignment algorithm in a similar fashion as~\cite{deflicker} by allowing a variable reference length. Since we want to consume the \textit{entire} cascaded sequence, we must have a final best alignment path on the last row (shown in a red box on figure~\ref{fig:lev}). However, we do \textit{not} want to consume all the causal sequence because it is likely to have more token due to the emission delay from the cascaded model. In other words, the deletions due to the time delay should be ignored. Thus, instead of using the bottom right cell as the end-path of our alignment, we sweep all the costs on the bottom row and use the cell with the lowest cost. In essence, the procedure allows us to discover how many causal tokens correspond to the time delay. We are now able to create a composite transcript: First, we copy \textit{all} the tokens from the cascaded model, namely ``\texttt{\textcolor{purple}{\_ro sa l ie \_how}}''. Then, we append the \textit{remaining} causal tokens, namely ``\texttt{\textcolor{blue}{\_are \_you}}'' to get a composite transcript ``\texttt{\textcolor{purple}{
\_ro sa l ie \_how} \textcolor{blue}{\_are \_you}}''.

Mathematically, if the cascaded transcript $x$ has length $m$ and the causal transcript $y$ has length $n$, then the Levenshtein algorithm will compute the cost $C(i, j)$ of aligning substrings $x[1, \ldots, i]$ and $y[1, \ldots, j]$ for $i \in [1, m]$ and $j \in [1, n]$. We then compute:
\begin{equation}
j^* = \argmin _{j \in [1, n]} \left(C(m, j) \right)
\end{equation}

and the composite transcript is (where $\oplus$ is a string concatenation):
\begin{equation}
z \triangleq x[1,\ldots,m] \oplus y[(j^*+1),\ldots,n]
\end{equation}

\subsection{Cropping partial results for non-quadratic growth}

Because we run the algorithm on streaming audio, we must make sure it doesn't delay the appearance of results. Further, on device applications require a low power consumption. One of the issue with the naive approach of section~\ref{sec:intuition} is that its run time is quadratic. For short audio segments, this is not a problem, but when decoding longer duration segments, the cost can become prohibitive.

\input{algo3}

This can be alleviated by capping the length of the sequences so that the shorter sequence is at most of length $M$. Thus, we remove the first $P = \max(\min(m, n) - M, 1)$ tokens of each string and only align $x[P, \ldots, m]$ against $y[P, \ldots, n]$. Then, the composite transcript is:
\begin{equation}
z \triangleq x[1, \dots, P] \oplus x[P + 1, \ldots, m] \oplus y[(P + j^*+1), \ldots, n]
\end{equation}

The intuition is that we only to have at most the last $M$ tokens of both transcripts to align them properly. Using the approach, we can make the algorithm linear without much reduction in its quality.

\subsection{Trimming cascaded input transcript}
In order to reduce the amount of flickering, we can trim the cascaded transcript. Instead of simply using the full $x[1, \dots,m]$ word pieces as an input, we can trim $T$ word pieces and only use $x[1, \ldots, \max(m-T, 1)]$. While this does not create an additional latency (since we still use the full causal transcript), this may cause the composite transcript to have lower quality.

\subsection{Alignment cost bailing and adding hysteresis}
If the causal and cascaded transcripts differ too much, alignment may not produce desirable results. Misalignment reduce the quality of the composite result and when they get corrected cause flickering. We can use the alignment cost to determine the distance between the two transcripts, and if it is too high, we simply do not create a composite transcript and keep the causal transcript. We defined two costs measurements. The first uses the full sequence, namely:
\begin{equation}
\rho _f \triangleq \frac{C(m, n)}{m}
\end{equation}

We can also limit ourselves to the end of the alignment and only consider the last $K$ tokens:

\begin{equation}
\rho _r (K) \triangleq \frac{ C \left(m, n\right) - C \left( \max(m - K, 0), \max(n - K, 0) \right) }{\min(K, m)}
\end{equation}

We only rewrite the transcript if the cost is below a settable threshold. However, if we had previously used a cascaded transcript to create a composite transcript, we fall back to the previously used one, so that we do not suddenly stop rewriting transcript.

\subsection{Overall algorithm}

We decompose the overall approach in three algorithms. In algorithm~\ref{algo:loop}, we show the entry point of our approach. The code keeps pulling for new results as long as there is still audio to be decoded. In case a partial result comes from the cascaded model, it is suppressed and we only store the text that would have been outputted. In case a partial result comes from the causal model, we rewrite it and show it to the user.

In algorithm~\ref{algo:hysteresis}, we show the hysteresis that allows to fall back on previously used cascaded partial results. We attempt to rewrite the causal partial using the latest cascaded partial. If the cost of rewriting such partial is low enough, we record the latest cascaded result and return the composite transcript. If the cost is too high, then we fall back to unconditionally rewriting the causal transcript using the latest cascaded result used. Thus, if the two models completely disagree, we fall back on the last time they agreed.

Finally, algorithm~\ref{algo:create_composite} shows how we create a composite partial result. It computes the alignment on a subset of the tokens and then outputs the composite tokens. The cost is computed during the reconstruction. For brevity, the algorithm for trimming of the tokens is omitted. In order to reduce the flickering of both the causal and cascaded models themselves, we applied the same deflickering algorithm as~\cite{deflicker} with $\alpha=0.2$ for our test model.

%% file: fig_lev.tex
\definecolor{arylideyellow}{rgb}{0.91, 0.84, 0.42}
\definecolor{verylightgray}{rgb}{0.92, 0.92, 0.92}

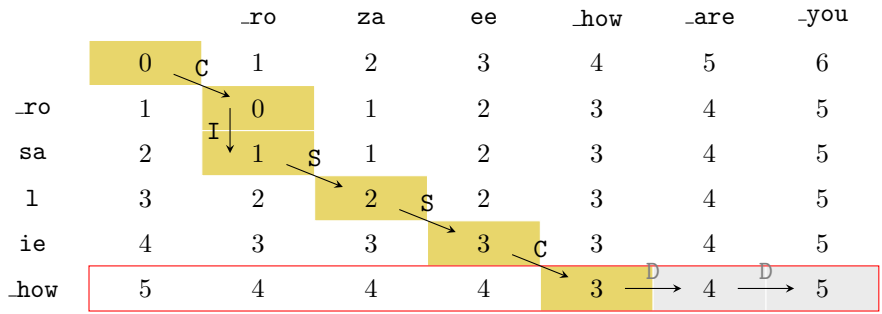
\begin{figure}[h!]
\begin{tikzpicture}[xscale=1.5,yscale=-0.60]
\draw (1, -1) node {\texttt{\_ro}};
\draw (2, -1) node {\texttt{za}};
\draw (3, -1) node {\texttt{ee}};
\draw (4, -1) node {\texttt{\_how}};
\draw (5, -1) node {\texttt{\_are}};
\draw (6, -1) node {\texttt{\_you}};

\draw (-1, 1) node {\texttt{\_ro}};
\draw (-1, 2) node {\texttt{sa}};
\draw (-1, 3) node {\texttt{l}};
\draw (-1, 4) node {\texttt{ie}};
\draw (-1, 5) node {\texttt{\_how}};

\draw[white,fill=arylideyellow] (-0.5, -0.5) rectangle (0.5, 0.5) {};  
\draw[white,fill=arylideyellow] ( 0.5,  0.5) rectangle (1.5, 1.5) {};  
\draw[white,fill=arylideyellow] ( 0.5,  1.5) rectangle (1.5, 2.5) {};  
\draw[white,fill=arylideyellow] ( 1.5,  2.5) rectangle (2.5, 3.5) {};  
\draw[white,fill=arylideyellow] ( 2.5,  3.5) rectangle (3.5, 4.5) {};  
\draw[white,fill=arylideyellow] ( 3.5,  4.5) rectangle (4.5, 5.5) {};  

\draw[white,fill=verylightgray] ( 4.5,  4.5) rectangle (5.5, 5.5) {};  
\draw[white,fill=verylightgray] ( 5.5,  4.5) rectangle (6.5, 5.5) {};  

\draw[red] (-0.5,  4.5) rectangle (6.5, 5.5);

\draw (0, 0) node {$0$};
\draw (1, 0) node {$1$};
\draw (2, 0) node {$2$};
\draw (3, 0) node {$3$};
\draw (4, 0) node {$4$};
\draw (5, 0) node {$5$};
\draw (6, 0) node {$6$};
\draw (0, 1) node {$1$};
\draw (1, 1) node {$0$};
\draw (2, 1) node {$1$};
\draw (3, 1) node {$2$};
\draw (4, 1) node {$3$};
\draw (5, 1) node {$4$};
\draw (6, 1) node {$5$};
\draw (0, 2) node {$2$};
\draw (1, 2) node {$1$};
\draw (2, 2) node {$1$};
\draw (3, 2) node {$2$};
\draw (4, 2) node {$3$};
\draw (5, 2) node {$4$};
\draw (6, 2) node {$5$};
\draw (0, 3) node {$3$};
\draw (1, 3) node {$2$};
\draw (2, 3) node {$2$};
\draw (3, 3) node {$2$};
\draw (4, 3) node {$3$};
\draw (5, 3) node {$4$};
\draw (6, 3) node {$5$};
\draw (0, 4) node {$4$};
\draw (1, 4) node {$3$};
\draw (2, 4) node {$3$};
\draw (3, 4) node {$3$};
\draw (4, 4) node {$3$};
\draw (5, 4) node {$4$};
\draw (6, 4) node {$5$};
\draw (0, 5) node {$5$};
\draw (1, 5) node {$4$};
\draw (2, 5) node {$4$};
\draw (3, 5) node {$4$};
\draw (4, 5) node {$3$};
\draw (5, 5) node {$4$};
\draw (6, 5) node {$5$};

\draw [-stealth](0.25, 0.25) -- (0.75, 0.75) node[above=0.01mm] [midway] {\texttt{\textbf{C}}};
\draw [-stealth](0.75, 1.00) -- (0.75, 2.00) node[left=0.01mm] [midway] {\texttt{\textbf{I}}};
\draw [-stealth](1.25, 2.25) -- (1.75, 2.75) node[above=0.01mm] [midway] {\texttt{\textbf{S}}};
\draw [-stealth](2.25, 3.25) -- (2.75, 3.75) node[above=0.01mm] [midway] {\texttt{\textbf{S}}};
\draw [-stealth](3.25, 4.25) -- (3.75, 4.75) node[above=0.01mm] [midway] {\texttt{\textbf{C}}};

\draw [-stealth](4.25, 5.00) -- (4.75, 5.00) node[above=0.01mm] [midway] {\texttt{\textcolor{gray}{D}}};
\draw [-stealth](5.25, 5.00) -- (5.75, 5.00) node[above=0.01mm] [midway] {\texttt{\textcolor{gray}{D}}};

\end{tikzpicture}
\caption{Example of Levenshtein alignment of two decoded sequences. The vertical is the partial result from the cascaded model $x[1, \ldots, m]$ with $m=5$, while the horizontal is the partial result from the causal model $y[1, \ldots, n]$ with $n=6$. The numbers correspond to the alignment cost $C(i, j)$ and the arrows the best path with C=correct, I=insertion, S=substitution, and D=deletion. The algorithm finds $j^* = 4$ that minimizes the cost of the last row, with $C(m, j^*)=3$.} 
\label{fig:lev}
\end{figure}

%% file: algo1.tex
\begin{algorithm}[h!]
\begin{algorithmic}
\scriptsize
\State{$\texttt{latest\_partial\_used\_for\_rewriting} \gets \texttt{""}$}
\State{$\texttt{latest\_cascaded\_partial} \gets \texttt{""}$}

\While{$\texttt{AudioStillAvailable()}$}
  \State{$\texttt{partial\_result}$ $\gets$ $\texttt{PullPartialResult()}$}
  
  \If{$\texttt{partial\_result.origin == CASCADED}$}
  
     \State{$\texttt{latest\_cascaded\_partial}$ $\gets$ $\texttt{partial\_result.text}$}
  \Else

     $\texttt{OutputPartial(RewriteResult(partial\_result.text))}$
  \EndIf 
\EndWhile

\end{algorithmic}
\caption{Streaming algorithm that recomposes results.}
\label{algo:loop}
\end{algorithm}

%% file: algo2.tex
\begin{algorithm}[h!]
\begin{algorithmic}
\scriptsize
\Procedure{RewriteResult}{$\texttt{causal\_partial}$}

\State{$\texttt{cost},\texttt{composite}$ $\gets$ $\texttt{CreateComposite(causal\_partial, latest\_cascaded\_partial)}$}

\If{$\texttt{cost} < \texttt{cost\_threshold}$}
  \State{$\texttt{latest\_partial\_used\_for\_rewriting} \gets \texttt{latest\_cascaded\_partial}$}

  \Return{$\texttt{composite}$}

\Else

  \State{$\_,\texttt{composite}$ $\gets$ $\texttt{CreateComposite(causal\_partial, latest\_partial\_used\_for\_rewriting)}$}

  \Return{$\texttt{composite}$}
\EndIf
\EndProcedure

\end{algorithmic}
\caption{Algorithm that attempts rewriting with a fall-back.}
\label{algo:hysteresis}
\end{algorithm}

%% file: algo3.tex
\begin{algorithm}[h!]
\begin{algorithmic}
\scriptsize
\Procedure{CreateComposite}{$\texttt{causal\_partial,cascaded\_partial}$}

\State{$\texttt{causal\_tok} \gets \texttt{Split(causal\_partial)}$}
\State{$\texttt{cascaded\_tok} \gets \texttt{Trim(Split(cascaded\_partial))}$}
\State{$\texttt{composite\_tok} \gets \texttt{\{\}}$}

\State{$\texttt{P} \gets \texttt{max(min(causal\_tok.size, cascaded\_tok.size) - M, 1)}$}

\State{$\texttt{codes} \gets \texttt{LevAlign(causal\_tok[P..], cascaded\_tok[P..])}$}

\For{$\texttt{j} \gets \texttt{1}$ to $\texttt{P}$}

    $\texttt{composite\_tok.append(cascaded\_tok[j])}$
\EndFor

\State{$\texttt{i} \gets \texttt{P}$}
\State{$\texttt{m} \gets 1$}
\State{$\texttt{cost} \gets 0.0$}

\For{$\texttt{code} \gets \texttt{codes}$}      
  \If{$\texttt{code == SUBSTITUTE || code == CORRECT}$}
  
    $\texttt{composite\_tok.append(cascaded\_tok[j])}$
    \If{$\texttt{j > cascaded\_tok.size - K}$}
      \If{$\texttt{code == SUBSTITUTE}$}
        \State{$\texttt{c} \gets \texttt{c + 1.0}$}
      \EndIf
      \State{$\texttt{m} \gets \texttt{m + 1}$}
    \EndIf
    \State{$\texttt{i, j} \gets \texttt{i + 1, j + 1}$}
  \EndIf
  \If{$\texttt{code == INSERT}$}
  
    $\texttt{composite\_tok.append(cascaded\_tok[j])}$
    \If{$\texttt{j > cascaded\_tok.size - K}$}
      \State{$\texttt{c} \gets \texttt{c + 1.0}$}
      \State{$\texttt{m} \gets \texttt{m + 1}$}
    \EndIf
    \State{$\texttt{j} \gets \texttt{j + 1}$}
  
  \EndIf
  \If{$\texttt{code == DELETE}$}
  
    \If{$\texttt{j > cascaded\_tok.size - K}$}
      \State{$\texttt{c} \gets \texttt{c + 1.0}$}
      \State{$\texttt{m} \gets \texttt{m + 1}$}
    \EndIf
    \State{$\texttt{i} \gets \texttt{i + 1}$}
  
  \EndIf
\EndFor

\While{$\texttt{i} < \texttt{causal\_tok.size}$}

    $\texttt{composite\_tok.append(causal\_tok[i])}$
\EndWhile

\Return{$\texttt{c / m, composite\_tok}$}

\EndProcedure

\end{algorithmic}
\caption{Algorithm that creates a composite result.}
\label{algo:create_composite}
\end{algorithm}

%% file: results.tex
\label{sec:results}

\input{result_table}

\subsection{Base model}

In order to evaluate the quality of our algorithm, we reused the model of~\cite{cascaded_e2e}.

It uses a streaming comformer-transducer where the encoder consists of 12 causal comformer layers~\cite{fast_emit}, each with 23 frames of left context, a self-attention with 8 heads, and a convolution kernel size of 15. We use an embedding prediction network with 2 previous labels as input~\cite{botros2021tied} and an embedding dimension of 320. The base model used fast emit~\cite{fast_emit} in order to reduce the latency of the output.

We used a frame rate of 30ms, but the encoder stacks two frames and then downsamples, the effective frame rate is 60ms. Further, because the cascaded model looks ahead 15 stacked frames, it means that its output roughly corresponds to what was said about 900ms in the past. We reused the model~\cite{multidomain_data} which was trained on 400k hours of multidomain data with a one-hot domain ID~\cite{fast_emit} indicating the type of utterance. All data are deidentified and the collection and handling abide by Google AI Principles~\cite{aiprinciples}.

\subsection{Results}

The results are shown on table~\ref{tab:results}. We do not report the regular WER as we verified experimentally that our algorithm indeed does not change the final results. We measure the UPWR, PWER, and partial latency on four test sets using the metrics described in section~\ref{sec:metrics_of_interest}. The test model is the result of a sweep of the parameters that gives a good balance between increased quality and flickering ($alpha=0.2$, $\rho_f=\infty$, $\rho_r=0.5$, $K=10$, $P=25$ and $T=1$). We can see that for all sets, the PWER is reduced significantly. The smaller reduction is for the voice search test set; this is explained by the average duration of the utterances. Since they are shorter in this test set, the 900ms delay of the cascaded model has greater impact. There is not as much time to rewrite the partial results. For the other test sets, the PWER is reduced by at least 10\%.

The picture for flickering is more mixed. Because we want to improve the quality of partial results, we must, by definition, change the decoded words and because we do not want to increase the latency, we must therefore change already decoded words and thus cause flicker. We do see a large increase in the flickering of the partial results (left sub-column of the UPWR section) for all sets, with a larger increase for test sets for longer utterances. However, the UPWR for the transition from the last partial result to the final result (middle sub-column) is significantly lower. This is explained by the fact that our algorithm pulls in the transition from causal results to cascaded results earlier. Therefore, at the end of the utterance, the partial results are closer to the final results. Overall, we see that the total UPWR (right sub-column) is lower.

As described in section~\ref{sec:metrics_of_interest}, only the change in latency metric is meaningful. For all tests sets, the change was less than 10ms, well below human perception. We also ran measurements on a Pixel6a phone and the time spent in each rewriting is on average below $0.1ms$ meaning that our algorithm has very low computation requirements and power consumption.

The reader can observe the effect of the algorithm in videos~\cite{supplementary_material1} that show the causal, cascaded, and merged partial results when decoding Librispeech~\cite{panayotov2015librispeech} utterances.

\subsection{Hyperparameter sweep analysis}

\begin{figure}[h!]
\includegraphics[width=1\textwidth]{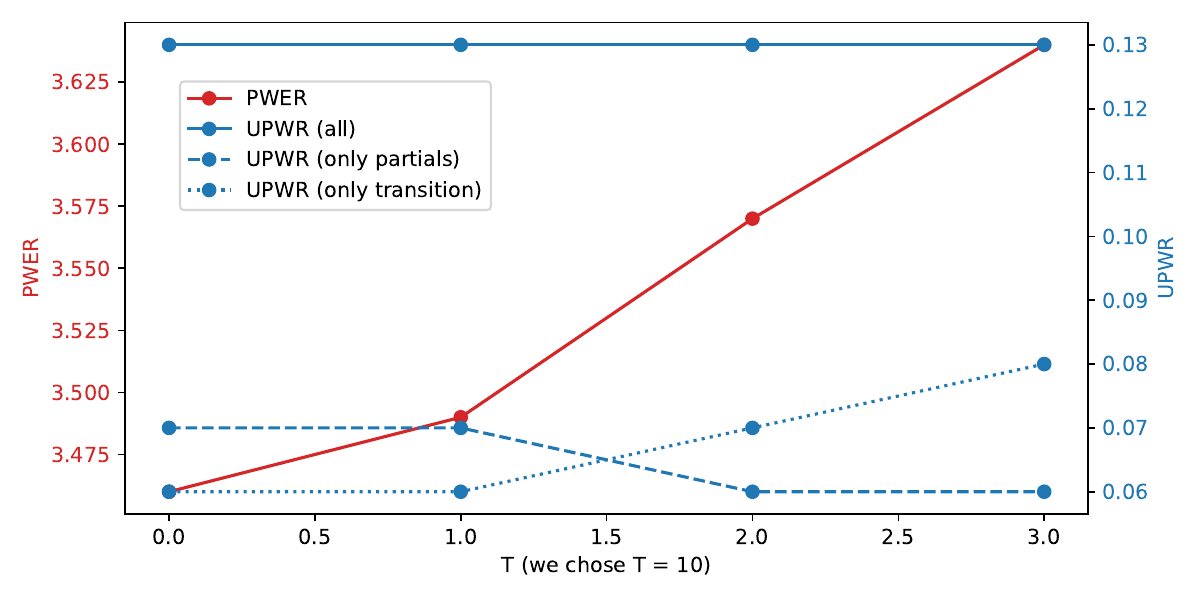}
\caption{PWER and UPWER as a function of $T$}
\centering
\label{fig:sweep_T}
\end{figure}

\begin{figure}[h!]
\includegraphics[width=1\textwidth]{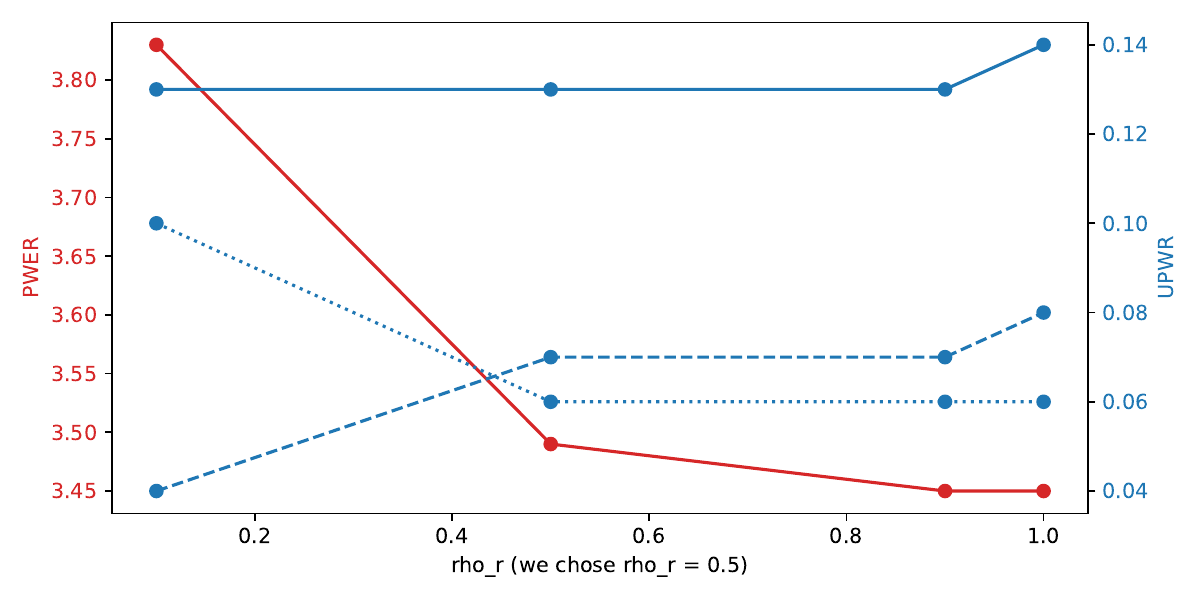}
\caption{PWER and UPWER as a function of $\rho_r$}
\centering
\label{fig:sweep_rho_r}
\end{figure}

\begin{figure}[h!]
\includegraphics[width=1\textwidth]{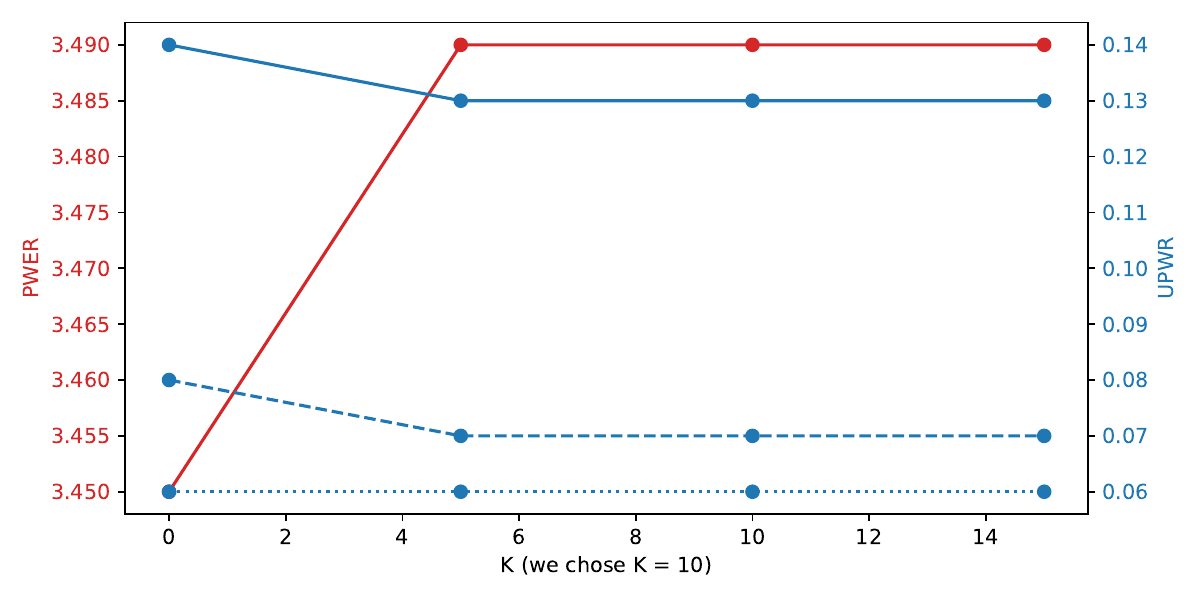}
\caption{PWER and UPWER as a function of $K$}
\centering
\label{fig:sweep_K}
\end{figure}

We plotted the PWER and the UPWR metrics according to the hyperparameter of the model for the dictation test set in figures~\ref{fig:sweep_T}, \ref{fig:sweep_rho_r}, and \ref{fig:sweep_K}. Note that the parameters we chose for the results of table~\ref{tab:results} are not the best according to the plots below.  This is because we chose a set of values that had acceptable results for the many test sets. This is a product judgment call, but we found that the parameters we chose work well for almost all our test sets. This suggests that the hyperparameters were not overfitted to the dictation test set.

We see on figure~\ref{fig:sweep_T} that as $T$ increases (the number of words we remove from the causal partial), the PWER goes up. This is expected because if we have fewer causal words (which are usually more correct), we cannot improve the rewritten transcript as much. As $T$ increases, we see that the UPWR for only the partials goes down (we rewrite less during decoding, thus we flicker less), and the UPWR for the transition goes up (by the time the final is shown, we have rewritten less and thus the transition has more flickering). Overall, the flickering is neutral.

We sweep the $\rho_r$ parameter in figure~\ref{fig:sweep_rho_r} where a lower value means that we want a smaller edit distance. When the parameter is 0.0, we recover the value of the base model of table~\ref{tab:results} (i.e. rewrite nothing). As it goes up, the PWER goes down because we rewrite more, but the flickering of the partial goes up (because we merge transcripts that are more different). As for the previous parameter, the transition UPWR goes down because we are closer to the final result.

We sweep the parameter $K$ in figure~\ref{fig:sweep_K}, which means how many words we use for the edit distance hysteresis. If we set it to 0, we recover the case where $\rho_r=1$. As K goes up, we are more strict when accepting rewrites, and thus the PWER goes up while the flickering goes down.

%% file: result_table.tex
\begin{table}[h!]
\begin{center}
\begin{tabular}{|ll|rrr|r|r|}
\hline
Test set & & & UPWR & & PWER \\
 &  & part. & trans. & all & \\
\hline
Dict- & Base & 0.04 & 0.12 & 0.15 & 3.89 \\
ation & Test & 0.07 & 0.06 & 0.13 & 3.49 \\
 & $\Delta$ & 75\% & -50\% & -13\% & -10\% \\
\hline
Voice & Base & 0.05 & 0.14 & 0.19 & 5.10 \\
Search & Test & 0.08 & 0.09 & 0.16 & 4.99  \\
 & $\Delta$ & 60\% & -36\% & -16\% & -2\% \\
\hline
TTS & Base& 0.03 & 0.51 & 0.54  & 3.64 \\
Audio & Test & 0.16 & 0.02 & 0.18 & 2.94 \\
book & $\Delta$ & 433\% & -96\% & -67\% & -19\%\\
\hline
Libri- & Base & 0.05 & 0.31 & 0.36 & 7.47 \\
speech & Test & 0.17 & 0.05 & 0.22 & 6.18 \\
 & $\Delta$ & 240\% & -84\% & -39\% & -17\% \\
\hline
Tele- & Base & 0.08 & 0.34 & 0.42 & 17.63 \\
phony & Test & 0.23 & 0.11 & 0.34 & 15.51 \\
 & $\Delta$ & 197\% & -68\% & -19\% & -12\% \\
\hline
\end{tabular}
\end{center}

\caption{Results of our proposed approach. As described in section~\ref{sec:metrics_of_interest}, we measure UPWR for all three subsets (for only the partials, only the transition, and for all results) and PWER for our base and test models. We also show the change ($\Delta$) either in percentage or milliseconds. The dictation test set are longer-form utterances corresponding to messages. The voice search test set are shorter queries seeking web results. We used text-to-speech (TTS) to synthesize long-form audio to produce long utterances. The librispeech test set is from~\cite{panayotov2015librispeech}. Finally, we also reported on a telephony set. All data was handled abiding by Google AI Principles~\cite{aiprinciples}.}

\label{tab:results}
\end{table}

%% file: conclusion.tex
Until recently~\cite{deflicker}, not much attention was paid to the quality of partial results of streaming ASR systems. It was assumed that improving final results would automatically improve partial results. Separately, more recent approaches for streaming decoding used multiple models~\cite{cascaded_e2e_original}. We presented an algorithm that doesn't require a re-architecturing of the model like in~\cite{jay2beamsearches} but instead relies on a simple, more general, rule to combine partials from multiple models. Doing so, we are able to significantly reduce partial word error rate and flickering without impacting partial latency. The computational cost of our approach is extremely low and therefore well-suited for on-device recognition. While we used a cascaded architecture to validate our approach, the proposed algorithm is applicable to any multi-decoder approach; For example, it can be applied to merging a stream of high quality but high latency partial results from a server recognizer into a stream of low-latency partial results from an on-device recognizer.

Our approach also opens a more flexible allocation of the model weights. We could re-allocate weights away from the causal model and towards the cascaded one. This could improve the quality of the final result, with limited impact on the partial results, precisely because we have rewriting.

Further work will focus on expanding the applicability of our approach. One area of further research is lowering the computational requirements of our systems by trading memory and computation away from the causal model towards the cascaded model without impacting the PWER too much. For example, the causal model could be run greedily with a single beam, or we could re-allocate the some weights to the cascaded model.

Another area of further research could be multi-stage and intermittent model. The approach described is easily expanded to 3 or more models, where the smallest model could be run more often, while the subsequent larger ones only as needed.